\documentclass[paper=a4, fontsize=11pt]{article}
\usepackage[T1]{fontenc}
\usepackage{fourier}

\usepackage[english]{babel}															
\usepackage{amsmath,amsfonts,amsthm} 
\usepackage[pdftex]{graphicx}	
\usepackage{url}
\usepackage{verbatim}
\usepackage{makecell}

\usepackage{sectsty}
\allsectionsfont{\centering \normalfont\scshape}

\usepackage{fancyhdr}
\numberwithin{equation}{section}		
\numberwithin{figure}{section}			
\numberwithin{table}{section}				

\newcommand*{\tabref}[1]{\tablename~\ref{#1}}
\newcommand*{\figref}[1]{\figurename~\ref{#1}}

\title{
Application of multiview techniques to NHANES dataset
}
\author{
		\normalfont 								\normalsize
        Aileme Omogbai\\[-3pt]		\normalsize
        Johns Hopkins University
}
\date{}

\begin{document}
\maketitle
\begin{abstract}
Disease prediction or classification using health datasets involve using well-known predictors associated with the disease as features for the models. This study considers multiple data components of an individual's health, using the relationship between variables to generate features that may improve the performance of disease classification models. 
In order to capture information from different aspects of the data, this project uses a multiview learning approach, using Canonical Correlation Analysis (CCA), a technique that finds projections with maximum correlations between two data views. Data categories collected from the NHANES survey (1999-2014) are used as views to learn the multiview representations. The usefulness of the representations is demonstrated by applying them as features in a Diabetes classification task.
\end{abstract}

%
\section{Introduction}
Research into disease-related health variables typically involve choosing health variables and conditions, and using statistical methods to study the strength of association of the variables with the condition~\cite{diabetes_assoc_a}. These are then used to confirm known or suspected relationships between the behavioural/health factors or disease conditions. 

There may be information about health status that may be gleaned by considering different aspects of an individual's data, and investigating possible relationships between the variables. Representations that capture these relationships can be useful in predicting presence or risk level of medical conditions. 
The National Health and Nutrition Examination Survey (NHANES) dataset provides data on health measurements, taken from survey participants, comprising different categories including demographics, laboratory tests and physical measurements. In this project, each of these categories are thought of as separate views of the participants' data. Canonical Correlation Analysis \cite{hoteling}, a multi-view learning technique, is used to learn representations that capture information from two different views.

The major consideration in this project is the use of the available NHANES data for multiview analysis. Multiview learning is useful when the data exists in partitions such that each partition(view) is a natural combination of the variables. The purpose of multiview learning is to exploit the redundancy or common information in the different views to improve performance or provide more insight into the data. The NHANES data fits naturally in a multiview scenario because the data is collected in categories pertinent to the overall health status of an individual. Demographics data, laboratory test results, or physical measurements are examples of natural partitions of the data that can be considered as individual views.

The features obtained from applying multiview techniques to the data optimize properties that help to incorporate information from the different views into a single representation. Therefore, these features may be useful for predicting or classifying certain health conditions of an individual. The multiview features may give additional information that will otherwise be lost by considering only pre-determined health indicators. This is investigated for the case of Diabetes classification, considering both known diabetes predictors and multiview features.

\section{NHANES}
The National Health and Nutrition Examination Survey (NHANES) is an ongoing survey designed to measure the health and nutritional status of the U.S. population. The survey referred to in this work is the continuous NHANES data, which was initiated in 1999, not to be confused with the  NHANES(I, II and III) surveys from previous years. The survey is conducted by the National Center for Health and Statistics (NCHS), which is a part of the Center for Disease Control and Prevention (CDC). The survey utilizes a nationally representative sample of the US noninstitutionalized, civilian population identified through a multistage probability sampling design. The sample for each year consists of about 5000 participants. 

\begin{figure}[!ht]
  \centering
    \includegraphics[width=.7\textwidth]{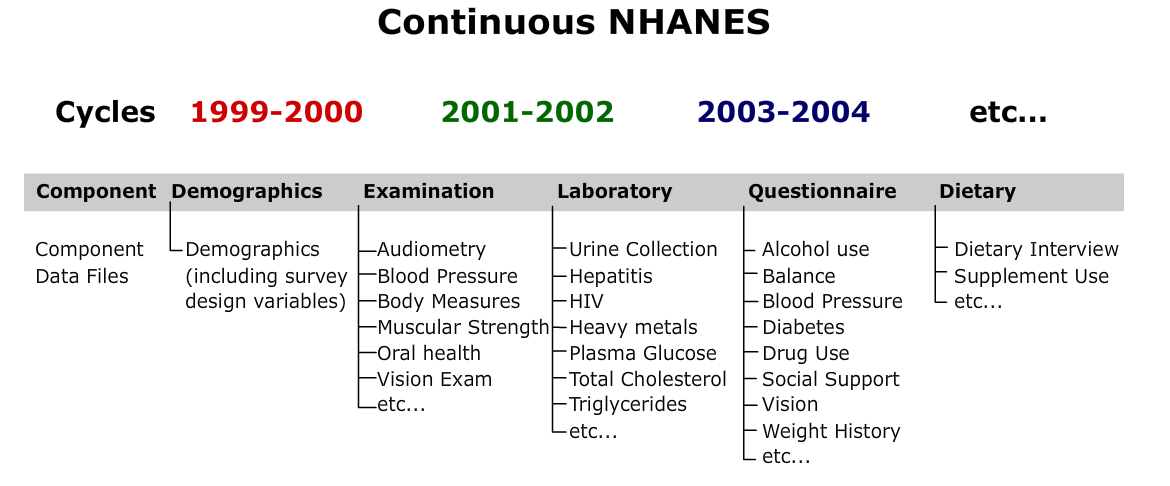}
  \caption{NHANES data structure~\cite{nhanes_data_structure_site}}
  \label{fig:data_Str}
\end{figure}

The survey consists of an interview section and a physical examination. The participants go through a comprehensive interview involving demographic, nutrition and health-related  behavioral questions. This includes questionnaires about participants' medical conditions, alcohol/tobacco use, and other medically pertinent details. The examination component consists of physical as well as physiological measurements, so it includes measures like height, weight, blood pressure and so on. Bodily fluids such as blood and urine are also collected, and the results of several laboratory tests are also released as part of the participants' data. The general structure of the data collected is shown in \figref{fig:data_Str}.

Information gotten from the NHANES data is used to determine the prevalence of major diseases and risk factors for diseases, assess nutritional status and its association with health promotion and disease prevention, and form the basis for national standards for such measurements as height, weight, and blood pressure \cite{about_nhanes}. The data is used to compile most of the statistics used to assess the health status of the country.

Though the dataset appears to be a rich source of health data for the population, there is a paucity of research work applying machine learning techniques to the dataset. A cursory Google Scholar~\cite{google_scholar} search shows little machine learning research on the NHANES data, especially considering the recent advancements in machine learning. A majority of the available literature involves investigation into the prevalence of disease conditions in certain population groups, or the association between specific variables and/or disease conditions. 

One of the problems with using the data is the difficulty in constructing a concrete dataset from the the survey data. Because the survey is conducted and reviewed yearly, the data available is inconsistent across the years span. Due to design considerations, variables may be added, removed or renamed for a new survey year, or the format might be changed totally. So compiling a complete dataset involves manually searching for variables across yearly releases, before dealing with data cleaning and missing data. 

\subsection{Dataset Compilation} 
\label{sub:dataset_compilation}
The NHANES dataset is released in the form of SAS XPORT files, where each file contains the results for a particular questionnaire, in a particular 2-year survey period. For example, demographics for the 1999-2000 period is contained in one file, and so on, for the different years available. At the time of this project, the data for 2015-2016 was incomplete, so the data used were for the surveys done between 1999 and 2014. The data files and related documentation are puclicly released and available at the NHANES website \cite{nhanes_data_website}. The documentation provides a description of each variable, the coding of its values, and the target individuals for that variable. 

There are problems with the dataset that make it difficult to compile the data from the different survey periods into a homogeneous, usable form. The data format for a particular category or questionnaire usually does not stay the same for different survey periods.
\begin{enumerate}
  \item Variables are, often times, named differently from previous years. It is also common to see variables recoded for different reasons, so special care must be taken to keep the values consistent across the years. For example, after 2010, the values for variable \textit{Country of Birth} in the demographics component was changed from \textit{United States / Mexico / Other Spanish / Other Non-spanish} to \textit{United States / Others}. 
  \item A variable may be discontinued, moved to a different file, or totally unavailable in a particular year. 
  \item Variables have different eligibility/target criteria (usually age, or the value of another variable), and some data are missing altogether.
\end{enumerate}
Compiling a dataset therefore involves manually checking the documentation for each data file and making appropriate cleaning decisions for each variable. It also limits the size of the dataset because variables may have to be removed to keep the data at a reasonable size.

\begin{table}[!ht]
  \begin{center}
    \begin{tabular}{|p{3cm}|p{9cm}|p{1cm}|}
    \hline
    \textbf{Category} & \textbf{Variables} & Size \\ \hline
    Demographics & Adult Education Level, Age, Birth Country, Citizen, Family Income, Family Income To Poverty Ratio, Gender, Age of Household Reference Person, Household Income, Birth Country of Household Reference Person, Education level of Household Reference Person, Gender of Household Reference Person, Marital Status of Household Reference Person, Household Size, Marital Status, Pregnant At Exam, Race/Ethnicity, Served In Military, Years In U.S & 37266\\ \hline
    Body Measures & Avg. BP Diastolic Pressure, Avg. BP Systolic Pressure, Body Mass Index, Height, Waist Circumference, Weight & 57160\\ \hline
    Laboratory (Tests) & White blood cell count (SI), Lymphocyte percent, Monocyte percent, Segmented neutrophils percent, Eosinophils percent, Basophils percent, Lymphocyte number, Monocyte number, Segmented neutrophils number, Eosinophils number, Basophils number, Red cell count SI, Hemoglobin (g/dL), Hematocrit , Mean cell volume (fL), Mean cell hemoglobin (pg), MCHC (g/dL), Red cell distribution width, Platelet count  SI, Mean platelet volume (fL) & 20990 \\ \hline
    Smoking & Start age, Number of days smoked(last 30 days), Average number smoked (last 30 days), Cigarette Filter, Cigarette Length, Nicotine content, How soon after waking & 5643 \\ \hline
    \end{tabular}
  \end{center}
  \caption{Post-cleaning variables for different data categories}
  \label{tab:variables}
\end{table}\tabref{tab:variables} shows the variables used in this project for most of the experiments carried out. More details about the compilation and data cleaning done to extract these variables, is available in the accompanying code documentation.
Also provided are scripts for easy access and usage of the dataset including:
\begin{enumerate}
  \item Downloading data component files for all available release cycles from the NHANES repository. 
  \item Data cleaning: removing invalid entries, harmonizing variable names, and merging data from the different release cycles.
  \item Exploratory data analysis on major data components e.g Demographics and Examination.
  \item Extraction of features and class variables for diabetes classification.
\end{enumerate}


\newpage
\section{Experimental Methods}

Analysis of the data begins with basic exploration, to provide a better understanding of the design choices made in the survey. From the data documentation, most variables have a target criteria, however basic demographic and body measure variables such as age, gender and height are taken for almost all the participants. 
\begin{figure}[!ht]
  \centering
    \includegraphics[width=.7\textwidth]{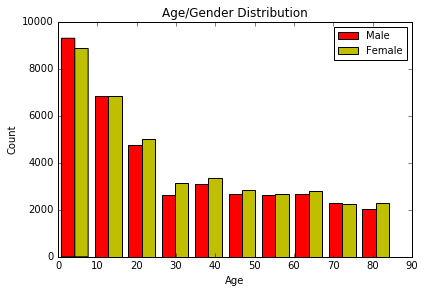}
  \caption{Age and Gender distribution in NHANES survey participants} 
  \label{fig:age_gender}
\end{figure}
\figref{fig:age_gender} shows the distribution of the ages of the survey individuals, and the gender breakdown in each age group. The histogram shows that the gender distribution is roughly even in the dataset. It also shows that the age distribution is right-skewed, with disproportionately more young particpants. The survey guidelines explain that this is an intentional, in order to have more data for children's health studies.

\begin{figure}[!ht]
  \centering
    \includegraphics[width=1\textwidth]{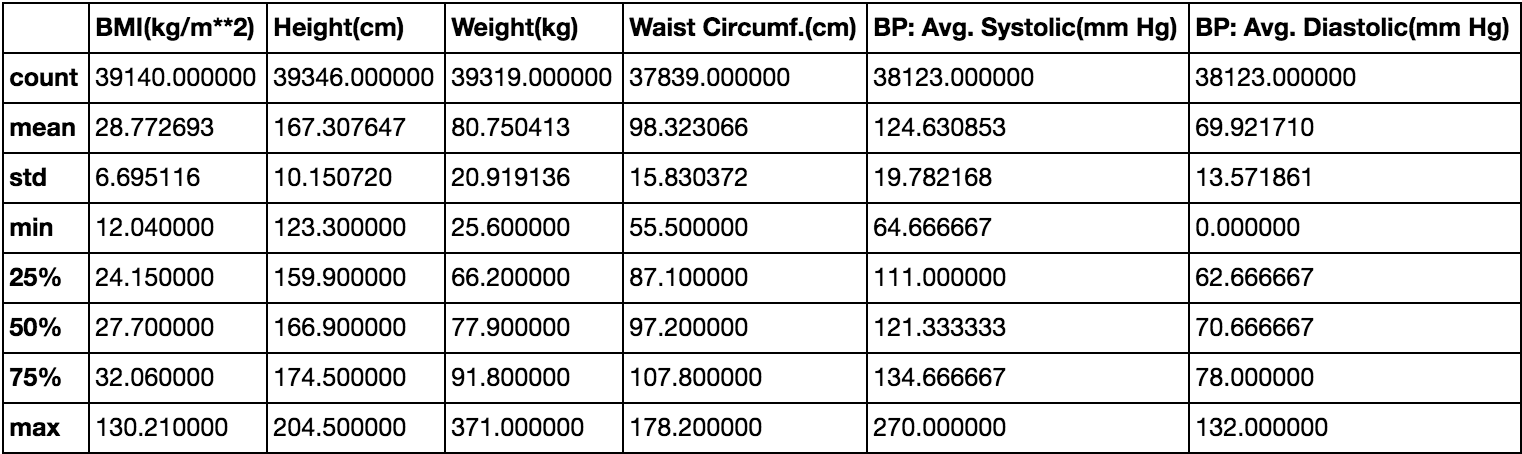}
  \caption{Summary of Body Measures data} 
  \label{fig:body_measures_descr}
\end{figure}
The dataset also contains data on physical and physiological body measurements. \figref{fig:body_measures_descr} shows summary statistics for the body measures data. The blood pressure readings are average of 3 different readings taken as part of the physical examination. For easy and consistent interpretation, only data for adults (age > 20) is used to generate the summary. 

\subsection{Principal Component Analysis}
Principal Component Analysis (PCA) is a useful technique for exploratory data analysis. Given a set of possibly correlated variables, PCA produces another set of variables called principal components, that are linear combinations of the original variables. The principal components are linearly uncorrelated and retain as much variance in the original dataset as possible. We typically focus on number of principal components that is much fewer than the number of original variables. 
Formally, consider a centered data matrix $X \in \mathbb{R}^{d \times n}$, where each of the $d$ columns represents an input variable, and each row represents a sample from the data population. PCA finds $k$ principal directions, $u_1, \ldots, u_k \in \mathbb{R}^d$, such that they solve the following optimization problem~\cite{arora2012stochastic,arora2013stochastic}:
\begin{align*}
  \max\limits_{u_i \in \mathbb{R}^d,  i \in 1,\ldots,k} & var(u_i^\top X) \\
  \text{subject to } & u_i^\top u_i = 1 \\
  & u_i^\top u_j = 0 \quad \forall i \neq j,\ \  i, j \in \{1,\ldots,k\} \\
  & var(u_1^\top X) \geq var(u_2^\top X) \geq \cdots \geq var(u_k^\top X) \geq 0\\
\end{align*}
The top-$k$ principal directions are given by the top-$k$ eigenvectors of the covariance matrix $\Sigma = \mathbb{E}[XX^\top]$. The projections of $X$ onto these principal directions, $y_i = u_i^\top X, i \in 1,\ldots,k$, are called the principal components of $X$.

PCA is widely used for dimensionality reduction because it provides a smaller number of variables, while retaining as much variance in the original data as possible. Furthermore, even though PCA is a non-convex optimization problem~, several tractable algorithms for PCA exist; see~\cite{Allerton,NIPS:13} for a survey. Examples of the applications of PCA include face detection in computer vision \cite{turk_eigenfaces} and gene expression \cite{yeung_pca_gene}. 

Analysis of the PCA loadings (weights of the original variables in the principal components) can be used to assess the level of importance of the variables in explaining the variation in the dataset. The projection of the data onto the principal directions give a new set of data features with lower dimensionality. The new features can be used for visualizing the dataset, or as features for a prediction/classification task.

\subsection{Canonical Correlation Analysis}

Canonical Correlation Analysis (CCA) extends the idea of PCA to the multiview scenario. CCA finds a new coordinate system for each view, that captures the correlation between two views. The new coordinates are linear combinations of the original, such that there is maximum correlation between them. The method was first introduced by Harold Hotelling in \cite{hoteling}.

Formally, given 
paired random vectors $(x, y) \in \mathbb{R}^{d_x} \times \mathbb{R}^{d_y}$ with an unknown joint distribution $\mathcal{D}$, CCA finds $u \in \mathbb{R}^{d_x} and v \in \mathbb{R}^{d_y}$, that solve the following optimization problem: 
\begin{align*}
  &\max \quad\text{correlation}\left(u^\top x, v^\top y\right) \\
  \implies &\max\quad \frac{\text{covar}\left(u^\top x, v^\top y\right)}{\sqrt{\text{var}\left(u^\top x\right)}\sqrt{\text{var}\left(v^\top y\right)}}\\
  \implies &\max\quad \frac{\mathbb{E}_{xy}[u^\top xy^\top v]}{\sqrt{\mathbb{E}_{x}[u^\top xx^\top u]}\sqrt{\mathbb{E}_{x}[v^\top yy^\top v]}}.
\end{align*}
Since the objective is not affected by scaling $u$ or $v$, $u$ and $v$ can be picked without loss of generality such that $\mathbb{E}_{x}[u^\top xx^\top u] = \mathbb{E}_{x}[v^\top yy^\top v] = 1$. This gives the following constrained optimization problem: 
\begin{align*}
  \max &\quad \mathbb{E}_{xy}[u^\top xy^\top v] \\
  \text{subject to} &\quad \mathbb{E}_{x}[u^\top xx^\top u] = 1, \mathbb{E}_{x}[v^\top yy^\top v] = 1
\end{align*}
Given $n$ data points $(x_i, y_i)$ sampled i.i.d. from $\mathcal{D}$, let $\widehat{\Sigma}_{xx}$ and $\widehat{\Sigma}_{yy}$ be the sample covariance matrices of $x$ and $y$, respectively, and let $\widehat{\Sigma}_{xy}$ be the sample cross-covariance matrix between $x$ and $y$. 
These are used as estimates of the population covariance and cross-covariance matrices in the optimization problem:
\begin{align*}
  \max &\quad u^\top \widehat{\Sigma}_{xy} v \\
  \text{subject to} &\quad u^\top \widehat{\Sigma}_{xx}u = 1, v^\top \widehat{\Sigma}_{yy} v = 1.
\end{align*}
The solution, $\widehat{u}$, to the above problem is given as the top eigenvector of $\widehat{\Sigma}_{xx}^{-1} \widehat{\Sigma}_{xy} \widehat{\Sigma}_{yy}^{-1} \widehat{\Sigma}_{yx}$, and solution $\widehat{v}$ is given as $\widehat{v} = \frac{\widehat{\Sigma}_{yy}^{-1} \widehat{\Sigma}_{yx}}{\sqrt{\lambda}}$, where $\lambda$ is the corrreponding eigenvalue. This gives the first pair of canonical components. Subsequent pairs are found in the same way, but with the constraint of being uncorrelated to previous ones. As with the first pair, the $i$th canonical component is the $i$th eigenvector of $\widehat{\Sigma}_{xx}^{-1} \widehat{\Sigma}_{xy} \widehat{\Sigma}_{yy}^{-1} \widehat{\Sigma}_{yx}$.

CCA is used to generate features that exploit the relationship between two sets of variables. CCA has been applied to tasks in natural language processing~\cite{haghighi_cca,dhillon_cca,rastogi2015multiview}, speech~\cite{Bharadwaj:12,arora_cca_b,andrew2013using,arora2014multi,wang2015deep}, social network analysis~\cite{benton2016learning}, and computer vision~\cite{hardoon_cca_2004, kidron_cca_2005}. There are several extensions of CCA based multiview learning techniques, including those based on kernel methods~\cite{arora_cca} and deep neural networks~\cite{andrew_dcca}. Much like PCA, even though non-convex, the CCA optimization problem, and related multiview learning problems using Partial Least Squares (PLS), can be solved efficiently~\cite{wang2015stochastic,arora2016stochastic}. 

The categories of data collected represent different parameters for assessing health status. They can be thought of as views, with each view representing different aspects of health, but for the same individual. In multiview learning, the aim is to analyse the relationship between the views in order to gain more insight into the common association, which is the individual's health in this case. Since each view is associated with the person's health, any intrinsic relationship between the measurements/questionnaires should be related to the health as well. This may give information that may not be captured by considering the views individually.

The loadings of the canonical variables (the weights of the original variables in each pair of canonical components) can be used to discover relationships between the variables in the different views. For each CCA component, the loadings provide information about which original variables are dominant in that component. The sign of the loadings on the variables also show the sign of the correlation between the variables.

Like PCA, the projection to the CCA components produces a new set of features for both data views. The number of components used can also be varied, depending on how much of the correlation objective is retained by the components. It should be noted that the learnt CCA components can be applied to new data or to data entries for which the second view is not available. Since the features represent information about both views, the features are useful for prediction/classification on a common association of the views. The CCA features are used on a classification task to measure any performance gains gotten from the multiview analysis.


\subsection{Predictive Analysis: Diabetes}
\label{sec:diabetes}
Diabetes is one of the major disease conditions intended to be studied using the NHANES survey data. Diabetes is the 7th leading cause of death in the U.S. affecting an estimated 29 million people, with 8 million of those undiagnosed \cite{ada}. Early screening and diagnosis is essential to effective prevention and management strategies \cite{diabetes_care}. As a result, many prediction models attempting to identify people at high risk of developing diabetes using machine learning techniques, have been developed \cite{diabetes_risk_calc, computational_diabetes_review}. 

\subsubsection{Diabetes classification} 
\label{sub:diabetes_paper}
Using the NHANES dataset, Yu et al.\cite{yu_wei_diabetes_svm} implement a Support Vector Machines (SVM) model for diabetes classification. The features used in the model are known diabetes risk predictors such as BMI, having hypertension, and family history of diabetes. 
\begin{table}[!ht]
  \begin{center}
    \begin{tabular}{ |p{2cm}|p{7cm}|p{1cm}|p{1.5cm}|p{1.5cm}|}
    \hline
    \thead{Diagnostic \\ Category} & \thead{Definition} & \thead{N} & \thead{Classification \\ Scheme I} & \thead{Classification \\ Scheme II}\\ \hline
    Diagnosed diabetes & Answered 'Yes' to the question "Have you ever been told by a doctor or health professionals that you have diabetes" & 1,266 & Cases & Excluded from analysis\\ \hline
    Undiagnosed diabetes & \makecell{Answered 'No' to the question \\ "Have you ever been told by a doctor or \\ health professionals that you have diabetes" \\ AND \\ Fasting plasma glucose level $\geq$ 126 mg/dl }  & 195 & Cases & Cases\\ \hline
    Pre-diabetes & Fasting plasma glucose level 100-125 mg/dl & 1,576 & Non-Cases & Cases \\ \hline
    No diabetes & Fasting plasma glucose level $\leq$ 100 mg/dl & 3277 & Non-Cases & Non-Cases \\ \hline
    \end{tabular}
  \end{center}
  \caption{Description of the data set used for the study in \cite{yu_wei_diabetes_svm}}
  \label{tab:schemes}
\end{table}

The NHANES survey includes a questionnaire on medical conditions, in which participants are asked about their past and present diagnosis and family history on major disease conditions. Participants are considered to have diagnosed diabetes if they answered 'Yes' to the question "Have you ever been told by a doctor or health professionals that you had diabetes?". This data is gotten from the diabetes questionnaire administered in the survey. The plasma glucose level of participants is measured as part of the laboratory tests component of the survey. \tabref{tab:schemes} describe the classification scheme and the criteria for allocating participants to the different classes. Only the NHANES data from 1999-2004 was used in \cite{yu_wei_diabetes_svm} for creating the models.

Since more data is available, this project attempts to recreate the models to run on all the data available. The features, with their location in the survey components, are described in \tabref{tab:features}. Some variables could not be located and were thus omitted from the model features.

\begin{table}[!ht]
  \begin{center}
    \begin{tabular}{|p{8cm}|p{6cm}|}
    \hline
    \textbf{Variable} & \textbf{Location in NHANES dataset} \\ \hline
    Family history of diabetes & Medical Conditions Questionnaire  \\ \hline
    Age, Gender, Race/ethnicity, Household income & Demograhics  \\ \hline
    Height, Weight, BMI, Waist circumference & Body Measures  \\ \hline
    Hypertension, Blood(Systolic and Diastolic) Pressure & Blood Pressure and Cholesterol Questionnaire  \\ \hline
    Alchohol use: Number of alcoholic drinks per day & Alcohol Use Questionnaire \\ \hline
    Smoking: Smoker or not, Number of cigarettes per day & Smoking - Cigarette use Questionnaire \\ \hline
    \end{tabular}
  \end{center}
  \caption{Features for diabetes classifier schemes}
  \label{tab:features}
\end{table}
As commonly used in the medical domain, the models are evaluated on test data using the Sensitivity, Specificity, Positive Predictive Value, Negative Predictive Value and the area under the curve (AUC) of the receiver operating characteristic (ROC) curve. However, cross-validation is done using the ROC-AUC, especially because of the inherent class imbalance since most of the individuals would not have the disease.

The variables used as features for the classification are variables that can be easily collected from individuals without any special medical procedures. The aim of the prediction model is to help identify individuals who have or are at risk of diabetes. Comprehensive tests could then be performed to properly diagnose and manage these cases. Since the NHANES data provides laboratory test results on the participants, that data could be leveraged to further improve the performance on the classification. For example, the canonical components produced by running CCA on demographics and laboratory data, produces features that are correlated with laboratory tests data. Those features can therefore be useful for classifying diabetes risk.

\section{Results and Discussion}

\subsection{Diabetes Classification} 
\label{sub:diabetes_classification}


In keeping with the work done in \cite{yu_wei_diabetes_svm}, SVMs are used to train the classification models for all the results produced. The optimal settings of the SVM hyperparameters (kernel, C, gamma) are estimated by doing 5-fold cross-validation on a grid search over a range of possible hyperparameter settings, with a 70/30 split for training and testing. 

Different models are trained using different features from the CCA results.
\begin{enumerate}
  \item REG: The model trained using the regular diabetes predictors.
  \item CCA-DL-$n$: Models trained using the projection of Demographics view on to the first $n$ CCA components on Demographics and Laboratory views. Only data points with both views available are used. 
  \item CCA-DL-$n$-ALL: Models trained using the projection of ALL Demographics data onto the CCA components from CCA-DL-$n$
  \item CCA-BL: Models trained using the projection of Body Measures view on to the CCA components on Demographics and Body Measures views.  Only data points with both views available are used. 
  \item CCA-BL-ALL: Models trained using the projection of ALL Body Measures data onto the CCA components from CCA-DL-$n$
  \item REG+[CCA]-$m$: Models trained using the best $m$ features of a CCA model features stacked with the REG model features. 
\end{enumerate}

For all the models, the RBF kernel gave the best results. 
Table \ref{tab:svm_class_results} shows the results for the different models. 
\begin{table}[!ht]
  \begin{center}
    \begin{tabular}{|p{3cm}|p{1cm}|p{2cm}|p{2cm}|p{1cm}|p{1cm}|p{1cm}|}
    \hline
    \textbf{Model} & \textbf{Data Size} & \textbf{Sensitivity} & \textbf{Specificity}& \textbf{PPV}& \textbf{NPV}& \textbf{AUC}\\ \hline
    REG & 17902 & 0.801 & 0.711 & 0.481 & 0.915 & \textbf{0.756} \\ \hline
    CCA-DL-15 & 13404 & 0.788 & 0.908 & 0.560 & 0.967 & \textbf{0.848}\\ \hline
    CCA-DL-15-ALL & 18867 & \textbf{0.818} & \textbf{0.589} & 0.415 & 0.901 & 0.704\\ \hline
    CCA-BL & 19659 & 0.820 & 0.674 & 0.226 & 0.970 & 0.747 \\ \hline 
    REG+[CCA-DL-15-ALL]-5 & 16580 & 0.791 & 0.701 & 0.471 & 0.909 & 0.747\\ \hline 
    REG+[CCA-DL-15-ALL]-10 & 16580 & 0.784 & 0.720 & 0.485 & 0.908 & 0.752\\ \hline 
    \end{tabular}
  \end{center}
  \caption{SVM Diabetes Classification results}
  \label{tab:svm_class_results}
\end{table}
To determine the best features for diabetes classification in this dataset, the REG features are ranked by their weights in the best linear SVM model~\cite{linear_svm_ranking}. The ranking of the features is:
\begin{enumerate}
  \item Waist circumference, Number of alcoholic drinks per day, Age, BMI, Weight, Height, Family history, Hypertension, Education Level, Gender, Income level, Number of cigarettes smoked per day, Race/Ethnicity, Smoking.
\end{enumerate}

The REG model scores are lower than those reported in \cite{yu_wei_diabetes_svm}, but are at par with the results in \cite{diabetes_risk_calc, computational_diabetes_review} for similar experiments. The most significant result comes from the model with the Demographics/Laboratory CCA features with ~85\% AUC. This highlights the importance of CCA in producing features that are correlated with the laboratory data. However, since the motivation is to have a model that can give some information about diabetes before the  actual test is run, it is necessary to measure the performance of the CCA features on data points not included in CCA training. This can be seen in the results for model $CCA-DL-15-ALL$, and the results show a drastic performance reduction in AUC from projecting all demographic data to the learned CCA components. An interesting detail for the result is that the main decline is in the specificity meaning that the model produces more false positives, but the sensitivity(recall) increases, beating even the REG model.

The same procedure is applied with the CCA features learnt using Laboratory data with Body Measures as views. This gives lower performance suggesting that the demographic variables may provide more information for classifying diabetes than the Body Measure variables.

Stacking the CCA features with the REG features gives worse results than the REG model, and so does not justify combining both set of features in that way. 

\section{Conclusions}
This project explores the application of multiview learning on NHANES survey data components to generate new representations of participants' health.
The study is motivated by the intuition that more insight can be gotten from considering different modalities of health data, and exploiting any correlations between them. The learned multiview features achieve promising results on a diabetes classification task.

A possible extensions to the project would be in constructing other views from the dataset and studying the suitability of the multiview features to relevant machine learning tasks. Other multiview techniques such as Deep Canonical Correlation Analysis \cite{andrew_dcca,wang2015deep,wang2016deep} could also be applied, with the learned representations compared. Future work will also include applying techniques that extend CCA to more than two views such as Generalized CCA~\cite{carroll_gcca, kettenring_gcca,rastogi2015multiview,arora2014multi}. 


\newpage


\begin{thebibliography}{h}
\bibitem{about_nhanes}
About the National Health and Nutrition Examination Survey,
\\\path{http://www.cdc.gov/nchs/nhanes/about_nhanes.htm}

\bibitem{nhanes_data_structure_site} 
Key Concepts About the Data Structure,
\\\path{http://www.cdc.gov/nchs/tutorials/NHANES/SurveyOrientation/DataStructureContents/Info1.htm}



\bibitem{wang15unsupervised}
Weiran Wang, Raman Arora, Karen Livescu, and Jeff Bilmes.
\newblock Unsupervised learning of acoustic features via deep canonical correlation analysis.
\newblock In \emph{International Conference on Acoustics, Speech and Signal
  Processing (ICASSP)}, 2015.


\bibitem{google_scholar}
Google Scholar \path {https://scholar.google.com/scholar?q=nhanes+}

\bibitem{andrew2013using}
Andrew, Galen and Arora, Raman and Bharadwaj, Sujeeth and Bilmes, Jeff and Hasegawa-Johnson, Mark and Livescu, Karen. Using articulatory measurements to learn better acoustic features. In Proc. Workshop on Speech Production in Automatic Speech Recognition, 2013. 


\bibitem{nhanes_data_website}
NHANES Datasets and Documentation
\\\path{http://www.cdc.gov/nchs/nhanes/nhanes_questionnaires.htm}

\bibitem{arora2013stochastic}
Raman Arora, Andrew Cotter, and Nathan Srebro.
\newblock Stochastic optimization for {PCA} using capped {MSG}.
\newblock In \emph{Advances in Neural Information Processing (NIPS)}, 2013.


\bibitem{hoteling}
Hotelling, H. (1935) The most predictable criterion. Journal of Educational Psychology 26, 139-142.



\bibitem{diabetes_assoc_a}
Nguyen, Ninh T., et al. "Association of hypertension, diabetes, dyslipidemia, and metabolic syndrome with obesity: findings from the National Health and Nutrition Examination Survey, 1999 to 2004." Journal of the American College of Surgeons 207.6 (2008): 928-934.

\bibitem{Allerton}
Raman Arora, Andrew Cotter, Karen Livescu, and Nathan Srebro.
\newblock Stochastic optimization for {PCA} and {PLS}.
\newblock In \emph{50th Annual Allerton Conference on Communication, Control,
  and Computing}, 2012.


\bibitem{diabetes_care}
Global Guideline for Type 2 Diabetes: recommendations for standard, comprehensive, and minimal care. Diabet Med. 2006, 23: 579-593. 10.1111/j.1464-5491.2006.01918.x.

\bibitem{arora_cca}
Arora, Raman and Livescu, Karen. Kernel CCA for multi-view learning of acoustic features using articulatory measurements. In MLSLP, 2012.


\bibitem{yu_wei_diabetes_svm}
Yu, Wei, et al. "Application of support vector machine modeling for prediction of common diseases: the case of diabetes and pre-diabetes." BMC Medical Informatics and Decision Making 10.1 (2010): 16.

\bibitem{wang2015stochastic}
Weiran Wang, Raman Arora, Karen Livescu, and Nathan Srebro.
\newblock Stochastic optimization for deep cca via nonlinear orthogonal
  iterations.
\newblock In \emph{2015 53rd Annual Allerton Conference on Communication,
  Control, and Computing (ALLERTON)}, pages 688--695. IEEE, 2015.


\bibitem{diabetes_risk_calc}
Heikes KE, Eddy DM, Arondekar B, Schlessinger L: Diabetes Risk Calculator: a simple tool for detecting undiagnosed diabetes and pre-diabetes. Diabetes Care. 2008, 31: 1040-1045. 10.2337/dc07-1150.

\bibitem{arora2014multi}
Arora, Raman and Livescu, Karen. Multi-view learning with supervision for transformed bottleneck features. In IEEE International Conference on Acoustics, Speech and Signal Processing (ICASSP), 2014.


\bibitem{computational_diabetes_review}
Shankaracharya, Odedra D, Samanta S, Vidyarthi AS. Computational Intelligence in Early Diabetes Diagnosis: A Review. The Review of Diabetic Studies : RDS. 2010;7(4):252-262. doi:10.1900/RDS.2010.7.252.

\bibitem{cichosz_diabetes}
Cichosz, Simon Lebech, et al. "Improved diabetes screening using an extended predictive feature search." Diabetes technology \& therapeutics 16.3 (2014): 166-171.

\bibitem{arora2016stochastic}
Raman Arora, Poorya Mianjy, and Teodor Marinov.
\newblock Stochastic optimization for multiview representation learning using
  partial least squares.
\newblock In \emph{Proceedings of The 33rd International Conference on Machine
  Learning}, pages 1786--1794, 2016.



\bibitem{turk_eigenfaces}
Turk, M.A. and Pentland, A.P., 1991, June. Face recognition using eigenfaces. In Computer Vision and Pattern Recognition, 1991. Proceedings CVPR'91., IEEE Computer Society Conference on (pp. 586-591). IEEE.

\bibitem{yeung_pca_gene}
Yeung, K.Y. and Ruzzo, W.L., 2001. Principal component analysis for clustering gene expression data. Bioinformatics, 17(9), pp.763-774.

\bibitem{Bharadwaj:12}
Sujeeth Bharadwaj, Raman Arora, Karen Livescu, and Mark Hasegawa-Johnson.
\newblock Multiview acoustic feature learning using articulatory measurements.
\newblock In \emph{Intl. Workshop on Stat. Machine Learning for Speech
  Recognition (IWSML)}, 2012.


\bibitem{haghighi_cca}
Haghighi, A., Liang, P., Berg-Kirkpatrick, T. and Klein, D., 2008, June. Learning Bilingual Lexicons from Monolingual Corpora. In ACL (Vol. 2008, pp. 771-779).

\bibitem{arora2012stochastic}
Raman Arora, Andrew Cotter, Karen Livescu, and Nathan Srebro.
\newblock Stochastic optimization for {PCA} and {PLS}.
\newblock In \emph{50th Annual Allerton Conference on Communication, Control,
  and Computing}, 2012.


\bibitem{dhillon_cca}
Dhillon, P., Foster, D.P. and Ungar, L.H., 2011. Multi-view learning of word embeddings via cca. In Advances in Neural Information Processing Systems (pp. 199-207).


\bibitem{arora_cca_b}
Arora, Raman and Livescu, Karen. Multi-view CCA-based acoustic features for phonetic recognition across speakers and domains. In ICASSP, 2013.



\bibitem{kidron_cca_2005}
Kidron, E., Schechner, Y.Y. and Elad, M., 2005, June. Pixels that sound. In 2005 IEEE Computer Society Conference on Computer Vision and Pattern Recognition (CVPR'05) (Vol. 1, pp. 88-95). IEEE.

\bibitem{wang2015deep}
Weiran Wang, Raman Arora, Karen Livescu, and Jeff Bilmes.
\newblock On deep multi-view representation learning.
\newblock In \emph{Proc. of the 32st Int. Conf. Machine Learning (ICML 2015)},
  pages 1083--1092, 2015.


\bibitem{hardoon_cca_2004}
Hardoon, D.R., Szedmak, S. and Shawe-Taylor, J., 2004. Canonical correlation analysis: An overview with application to learning methods. Neural computation, 16(12), pp.2639-2664.

\bibitem{rastogi2015multiview}
Pushpendre Rastogi, Benjamin Van~Durme, and Raman Arora.
\newblock Multiview lsa: Representation learning via generalized cca.
\newblock In \emph{Proceedings of NAACL}, 2015.


\bibitem{kettenring_gcca}
Kettenring, J.R., 1971. Canonical analysis of several sets of variables. Biometrika, 58(3), pp.433-451.



\bibitem{carroll_gcca}
Carroll, J.D. 1968. Generalization of canonical correlation analysis to three or more sets of variables. Proceedings of the American psychological association (pp. 227–228)

\bibitem{benton2016learning}
Adrian Benton, Raman Arora, and Mark Dredze.
\newblock Learning multiview embeddings of twitter users.
\newblock In \emph{Proceedings of ACL}. ACL, 2016.


\bibitem{van-de-veldev_gcca}
Van De Velden, M. and Bijmolt, T.H., 2006. Generalized canonical correlation analysis of matrices with missing rows: a simulation study. Psychometrika, 71(2), pp.323-331.

  
  \bibitem{wang2016deep}
Weiran Wang, Raman Arora, Karen Livescu, and Jeff Bilmes.
\newblock On deep multi-view representation learning: Objectives and
  optimization.
\newblock \emph{arXiv preprint arXiv:1602.01024}, 2016.


\bibitem{linear_svm_ranking}
Chang, Yin-Wen, and Chih-Jen Lin. "Feature ranking using linear SVM." WCCI Causation and Prediction Challenge. 2008.

\bibitem{andrew_dcca}
Andrew, G., Arora, R., Bilmes, J.A. and Livescu, K., 2013, May. Deep Canonical Correlation Analysis. In ICML (3) (pp. 1247-1255).




  
\bibitem{ada}
American Diabetes Association. \path{http://www.diabetes.org}

\bibitem{NIPS:13}
Raman Arora, Andrew Cotter, and Nathan Srebro.
\newblock Stochastic optimization for {PCA} using capped {MSG}.
\newblock In \emph{Advances in Neural Information Processing (NIPS)}, 2013.


\end{thebibliography}
\end{document}